# Approche Multicritère pour le Problème de Ramassage et de Livraison avec Fenêtres de Temps à Plusieurs Véhicules

**I. Harbaoui Dridi**[(1),(2)]    **R. Kammarti**[(1),(2)]    **M. Ksouri**[(2)]    **P. Borne**[(1)]
imenharbaoui@gmail.com    kammarti.ryan@planet.tn    Mekki.Ksouri@insat.rnu.tn    p.borne@ec-lille.fr

[(1)] LAGIS : Ecole Centrale de Lille, Villeneuve d'Ascq, FRANCE
[(2)] LACS : Ecole Nationale des Ingénieurs de Tunis, Tunis - Belvédère. TUNISIE

**Résumé :** De nos jours, le problème de transport de marchandise occupe une place importante dans la vie économique des sociétés modernes. Le problème de ramassage et de livraison (pick-up and delivery problem) est l'un des problèmes dont une grande partie des recherches s'y est intéressée.
Il s'agit de déterminer un circuit de plusieurs véhicules, de façon à servir à coût minimal un ensemble de clients répartis dans un réseau, satisfaisant certaines contraintes relatives aux véhicules, à leurs capacités et à des précédences entre les nœuds. Dans cet article, nous présentons un état de l'art du VRP et du PDPTW, en proposant par la suite une approche multicritère basée sur les algorithmes génétiques qui permet de minimiser le compromis entre le nombre de véhicules, la somme des retards et le coût total de transport. Et ceci, en traitant le cas où un client peut posséder plusieurs fournisseurs et un fournisseur peut avoir plusieurs clients.
**Mots clés :** PDPTW, Evaluation multicritère, Algorithmes génétiques.

## 1. Introduction

Avec les contraintes temporelles et économiques qu'impliquent le problème de ramassage et de livraison, sa résolution devient très difficile, nécessitant l'utilisation d'outils issus de disciplines différentes (productique, informatique, optimisation combinatoire, etc.). En effet, les processus issus des systèmes de transports et de l'ordonnancement sont de plus en plus complexes, par leurs dimensions importantes, par la nature de leurs relations dynamiques, et par la multiplicité des contraintes auxquelles ils sont soumis.

Plusieurs recherches se sont principalement orientées vers la résolution du problème de tournées de véhicules (**VRP** : Vehicle Routing Problem). Ce dernier s'agit d'un problème d'optimisation de tournées de véhicules devant satisfaire des demandes de transport.

Une autre grande partie des recherches s'est intéressée à une importante variante du VRP qui est le **PDPTW** avec des contraintes de capacité sur les véhicules.

Le PDPTW qui est le problème de collecte et distribution à fenêtres de temps (Pickup and Delivery Problem with Time Windows), se divise en deux : **1-PDPTW** (à un véhicule) et **m-PDPTW** (à plusieurs véhicules).

Nous allons nous intéresser au problème de collecte et de distribution à fenêtre de temps, à plusieurs véhicules. Notre but est de concevoir une approche multicritère pour la résolution du m-PDPTW basée sur les algorithmes génétiques et qui permet de minimiser le compromis entre le nombre de véhicules, la somme des retards et le coût total de transport. Et ceci, en traitant le cas où un client peut posséder plusieurs fournisseurs et un fournisseur peut avoir plusieurs clients.

## 2. Etat de l'art

Le problème général de construction de tournées de véhicules est connu sous le nom de *Vehicle Routing Problem* (VRP) et représente un problème d'optimisation combinatoire multi-objectif qui a fait l'objet de nombreux travaux et de nombreuses variantes dans la littérature. Il appartient à la catégorie NP-difficile. [1] [Christofides, N et al] [2] [Lenstra, J et al]

Des métaheuristiques ont été aussi appliquées pour la résolution des problèmes de tournée de véhicules. Parmi ces méthodes nous pouvons citer les algorithmes de colonies de fourmis, qui ont été utilisés par [3] [Montamenni, R et al] pour la résolution du DVRP (Dynamic Vehicle Routing Problem). Ce dernier traite une composante dynamique dont elle s'exprime par les apparitions de nouveaux clients, de nouvelles demandes, ou de pannes des véhicules de transport.

Le principe du VRP est, étant donné un dépôt D et un ensemble de commandes de clients C = {c1, . . ., cn}, de construire un ensemble de tournées, pour un nombre fini de véhicules, commençant et finissant à un dépôt. Dans ces tournées, un client doit être desservi une seule fois par un seul véhicule et la capacité de transport d'un véhicule pour une tournée ne doit pas être dépassée. [4] [Nabaa, M et al]

Le problème de VRP avec des collectes et livraison de marchandises à fenêtres de temps correspond au PDPTW (*Pickup & Delivery Problem with Time Windows* ou Problème de Ramassage & Livraison avec Fenêtres de temps).

### 3. Le PDPTW: Pickup and Delivery Problem with Time Windows

Le PDPTW est une variante du VRPTW où en plus de l'existence des contraintes temporelles, ce problème implique un ensemble de clients et un ensemble de fournisseurs géographiquement localisés. Chaque tournée doit également satisfaire les contraintes de précédence pour garantir qu'un client ne doit pas être visité avant son fournisseur. [5] [Psaraftis, H.N]

Cette variante de problème se divise en deux catégories : le 1-PDPTW (avec un seul véhicule) et le m-PDPTW (à plusieurs véhicules).

Une approche de programmation dynamique pour résoudre le 1-PDP sans et avec fenêtres de temps a été développée par [6] [H.N. Psaraftis] en considérant comme fonction objectif la minimisation d'une pondération de la durée totale de la tournée et de la non-satisfaction des clients.

[7] [Jih, W et al] ont développé une approche basée sur les algorithmes génétiques hybrides pour résoudre le 1-PDPTW, ayant pour objectif de minimiser une combinaison du coût total et de la somme des temps d'attente. L'algorithme développé utilise quatre différents opérateurs de croisement ainsi que trois opérateurs de mutation.

Dans d'autres travaux, un algorithme génétique a été développé par [8] [Velasco, N et al] pour résoudre le 1-PDP bi-objectif dans lequel la durée totale des tournées doit être minimisée tout en satisfaisant en priorité les demandes les plus urgentes. Dans cette littérature, la méthode proposée, pour résoudre ce problème est inspirée d'un algorithme appelé Non dominated Sorting Algorithm (NSGA-II).

D'autres travaux traitent le 1-PDPTW, en minimisant le compromis entre la distance totale parcourue, le temps total d'attente et le retard total et ce en utilisant un algorithme évolutionniste avec des opérateurs génétiques spéciaux, la recherche taboue et la Pareto optimalité pour fournir un ensemble de solutions viables. [9] [Kammarti, R et al] Ces travaux ont été étendus, et ce en proposant une nouvelle approche basée sur l'utilisation de bornes inférieures pour l'évaluation des solutions et de leur qualité, minimisant le compromis entre la distance totale parcourue et la somme des retards. [10] [Kammarti, R et al]

En ce qui concerne le m-PDPTW, [11] [Sol, M et al] ont proposé un algorithme de *branch and price* pour résoudre le m-PDPTW et ce en minimisant le nombre de véhicules nécessaires pour satisfaire toutes les demandes de transport et la distance totale parcourue.

[12] [Quan, L et al] ont présenté une heuristique de construction basée sur le principe d'insertion ayant pour fonction objectif, la minimisation du coût total, incluant les coûts fixes des véhicules et les frais de déplacement qui sont proportionnels à la distance de déplacement.

Une nouvelle métaheuristique se basant sur un algorithme tabou intégré dans un recuit simulé, a été développée par [13] [Li, H et al] pour résoudre le m-PDPTW, ceci en recherchant, à partir de la meilleure solution courante s'il y a amélioration pendant plusieurs itérations.

[14] [Li, H et al] ont développé une méthode appelée « *Squeaky wheel* » pour résoudre le m-PDPTW avec une recherche locale.

### 4. Modèle mathématique adopté

Notre objectif est de servir toutes les demandes des clients, en minimisant le compromis entre le nombre de véhicules, la somme des retards et le coût total de transport. Notre problème est caractérisé par les paramètres suivants :

- N : Ensemble des nœuds clients, fournisseurs et dépôt,
- N' : Ensemble des nœuds clients, fournisseurs,
- N+ : Ensemble des nœuds fournisseurs,
- N− : Ensemble des nœuds clients,
- K : Nombre de véhicules,
- $d_{ij}$ : distance euclidienne entre le nœud i et le nœud j, si $d_{ij} = \infty$ alors le chemin entre i et j n'existe pas (impasse, rue piétonne,…)
- $t_{ijk}$ : temps mis par le véhicule k pour aller du nœud i au nœud j,
- $[e_i, l_i]$ : fenêtre de temps du nœud i,
- $s_i$ : temps d'arrêt au nœud i,
- $q_i$ : quantité à traiter au nœud i. Si $q_i > 0$, le nœud est fournisseur ; Si $q_i < 0$, le nœud est un client et si $q_i = 0$ alors le nœud a été servi.
- $Q_k$ : capacité du véhicule k,
- i = 0..N : indice des nœuds prédécesseurs,
- j = 0..N : indice des nœuds successeurs,
- k = 1..K : indice des véhicules,
- $Xijk = \begin{cases} 1 \text{ Si le véhicule k voyage du noeud i vers le noeud j} \\ 0 \text{ Sinon} \end{cases}$

- $A_i$ : temps d'arrivée au nœud i,
- $D_i$ : temps de départ du nœud i,
- $y_{ik}$ : quantité présente dans le véhicule k visitant le nœud i,
- $C_k$ : coût de transport associé au véhicule k,
- Un nœud (fournisseur ou client) peut être servi par un ou plusieurs véhicules.

- Un Client peut avoir un ou plusieurs fournisseurs et un fournisseur peut avoir un ou plusieurs clients.
- Il y a un seul dépôt,
- Les contraintes de capacité doivent être respectées,
- Les contraintes de temps sont rigides concernant l'heure d'arrivée,
- Chaque véhicule commence le trajet du dépôt et y retourne à la fin,
- Un véhicule reste à l'arrêt à un nœud le temps nécessaire pour le traitement de la demande.
- Si un véhicule arrive au nœud i avant la date $e_i$ de début de sa fenêtre, il attend.

La fonction à minimiser est donnée comme suit :

$$Minimiser\ f = \begin{cases} \lambda_1 K + \\ \lambda_2 \sum_{i \in N} \sum_{j \in N} max(0, D_i - l_i) + \\ \lambda_3 \sum_{i \in N} \sum_{j \in N} \sum_{k \in K} C_k d_{ijk} X_{ijk} \end{cases} \quad (1)$$

Avec:

$$\sum_{i=1}^{N} \sum_{k=1}^{K} x_{ijk} \geq 1,\ j = 2,...N \quad (2)$$

$$\sum_{j=1}^{N} \sum_{k=1}^{K} x_{ijk} \geq 1,\ i = 2,...N \quad (3)$$

$$\sum_{i \in N} X_{i0k} = 1, \forall k \in K \quad (4)$$

$$\sum_{j \in N} X_{0jk} = 1, \forall k \in K \quad (5)$$

$$\sum_{i \in N} X_{iuk} - \sum_{j \in N} X_{ujk} = 0, \forall k \in K, \forall u \in N \quad (6)$$

$$X_{ijk} = 1 \Rightarrow y_{jk} = y_{ik} + q_i, \forall i,j \in N; \forall k \in K \quad (7)$$

$$y_{0k} = 0, \forall k \in K \quad (8)$$

$$Q_k \geq y_{ik} \geq 0, \forall i \in N; \forall k \in K \quad (9)$$

$$D_w \leq D_v, \forall i \in N; \forall w \in N_i^+; \forall v \in N_i^- \quad (10)$$

$$D_0 = 0 \quad (11)$$

$$X_{ijk} = 1 \Rightarrow e_i \leq A_i \leq l_i, \forall i,j \in N; \forall k \in K \quad (12)$$

$$X_{ijk} = 1 \Rightarrow e_i \leq A_i + s_i \leq l_i, \forall i,j \in N; \forall k \in K; s_i \neq 0 \quad (13)$$

$$X_{ijk} = 1 \Rightarrow D_i + t_{ijk} \leq (l_j - s_j), \forall i,j \in N; \forall k \in K \quad (14)$$

Où $\lambda_i$ sont des coefficients de pondération et de mise à l'échelle.

Les équations (2) et (3) assurent qu'un nœud peut être servi une ou plusieurs fois par un ou plusieurs véhicules.
Les équations (4) et (5) assurent le non dépassement de la disponibilité d'un véhicule.
Un véhicule ne sort du dépôt et n'y revient qu'une seule fois.
L'équation (6) assure la continuité d'une tournée par un véhicule : le nœud visité doit impérativement être quitté.
Les équations (7), (8) et (9) assurent le non dépassement de la capacité de transport d'un véhicule.
Les équations (10) et (11) assurent le respect des précédences.
Les équations (12), (13) et (14) assurent le respect des fenêtres de temps.

### 5. Algorithme génétique pour l'optimisation multicritère du m-PDPTW

Le principe de différentes opérations génétiques telles que le codage du chromosome, la génération des populations ainsi que les procédures de correction de capacité et de précédence sont détaillés dans notre travail [15] [I. Harbaoui Dridi et al].

#### 5.1 Evaluation multicritère

Un problème multi objectif ou multicritère est défini comme un problème d'optimisation vectoriel, dont on cherche à optimiser plusieurs composantes d'un vecteur fonction coût.

#### 5.1.1 Définition et optimalité au sens de Pareto

Un problème multicritère $P$ constitué de $n$ variables, $m$ contraintes d'inégalités, $p$ contraintes d'égalités et $k$ critères peut être formulé de la manière suivante

$$P \Rightarrow \begin{cases} min\ f(x) = [f_1, f_2, f_3, ...... f_k(x)] \\ g_i(x) \leq 0\ \ i = 0...m \\ g_j(x) = 0\ \ j = 0...p \end{cases} \quad (15)$$

$x$ étant un vecteur solution $x = (x_1,...x_n)$ d'un univers $E$ de dimension $n$, appelé *espace de décision* (*ou espace des paramètres*). L'espace défini par $f = (f_1, f_2...f_k)$ noté $U$, est appelé *espace des critères* (*ou espace des objectifs*). La fonction d'évaluation $F : E \rightarrow U$, d'un problème multicritère, fait donc correspondre un vecteur $y( y = y_1,...y_k )$ de dimension $k$ tel que $y_k = f_k( x )$ à tout vecteur $x$ de dimension $n$ de l'espace $E$.

Il s'agit donc de trouver des solutions représentant un compromis possible entre les critères. Dans ce domaine, le concept de Pareto - optimalité introduit par l'économiste V. Pareto au XIXème siècle est fréquemment utilisé [16] [Pareto].

Au XIXème siècle, V. Pareto, formule le concept suivant : dans un problème multicritère, il existe un équilibre tel que l'on ne peut pas améliorer un critère sans détériorer au moins un des autres. Cet équilibre a été appelé optimum de Pareto. Un point est dit Pareto optimal s'il n'est dominé par aucun autre point appartenant à l'espace des solutions. Ces points sont également appelés solutions *non inférieures* ou *non dominées*.

Un point $X \in E$ domine $Y \in E$ si est seulement si :

$$\begin{cases} \forall i \in E, \quad f_i( x ) \leq f_i( y ) \\ et \ \exists j, \ \textbf{tel que} f_i( x ) < f_j( y ) \end{cases} \quad (16)$$

La figure 1 illustre un exemple où l'on cherche à minimiser $f_1$ et $f_2$. Les points 1,3 et 5 ne sont pas dominés. Par contre le point 2 est dominé par le point 3, et le point 4 est dominé par le point 5.

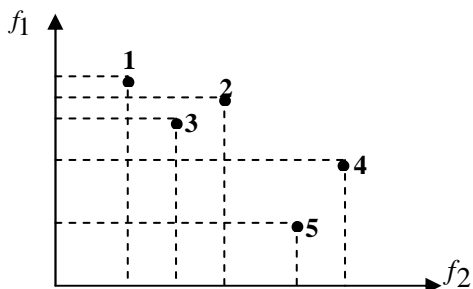

**Figure 1 : Exemple de dominance**

### 5.1.2 Méthode d'agrégation

Dans la résolution de PMO (Problème Multi objectifs), plusieurs méthodes traditionnelles transforment le PMO en un problème mono-objectif. Parmi ces méthodes on trouve la méthode d'agrégation. C'est l'une des premières méthodes utilisée pour la génération de solutions Pareto optimales. Elle consiste à transformer le problème (PMO) en un problème $( PMO_\lambda )$ qui revient à combiner les différentes fonctions coût $f_i$ du problème en une seule fonction objectif $F$ généralement de façon linéaire [17] [Hwang and Masud] :

$$F( x ) = \sum_{i=1}^{n} \lambda_i f_i( x ) \quad (17)$$

Où $\lambda_i$ sont des coefficients de pondération et de mise à l'échelle.

Le fonctionnement de la méthode d'agrégation (Figure 2) s'agit de fixer un vecteur poids c.à.d. trouver un hyper-plan dans l'espace objectif (une droite pour un problème bi-critères) avec une orientation fixée. La solution Pareto optimale est le point où l'hyper-plan possède une tangente commune avec l'espace réalisable.

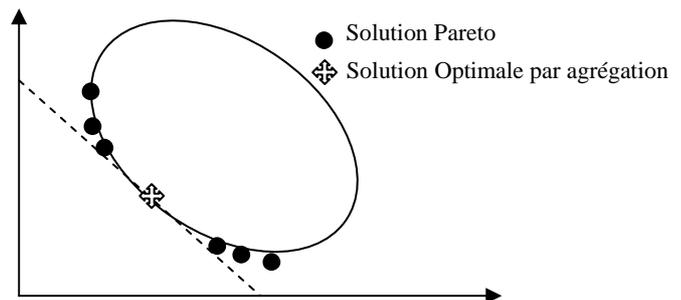

**Figure 2 : Exemple d'optimisation par agrégation à deux objectifs**

L'avantage de la méthode d'agrégation est la production d'une seule solution et ne nécessitent donc pas d'interaction avec le décideur.

### 5.2 Simulation et résultats

Le tableau 1 représente les paramètres caractérisant notre problème.

**Tableau 1 : Paramètres du problème**

| Id | X | Y | q | e | l | s | Succ | Pred |
|---|---|---|---|---|---|---|---|---|
| 0 | 0 | 0 | 0 | 0 | 200 | 0 | 0 | 0 |
| 1 | 57 | 26 | -20 | 41 | 67 | 3 | 0 | 2,9 |
| 2 | 60 | 57 | 20 | 34 | 100 | 11 | 1 | 0 |
| 3 | 24 | 77 | -20 | 69 | 124 | 2 | 0 | 4 |
| 4 | 22 | 17 | 20 | 78 | 158 | 13 | 3,7 | 0 |
| 5 | 87 | 1 | 20 | 62 | 64 | 13 | 6 | 0 |
| 6 | 50 | 60 | -20 | 5 | 145 | 4 | 0 | 5 |
| 7 | 28 | 93 | -20 | 27 | 81 | 1 | 0 | 4 |
| 8 | 84 | 5 | 20 | 61 | 91 | 11 | 12 | 0 |
| 9 | 40 | 11 | 20 | 95 | 142 | 13 | 1 | 0 |
| 10 | 4 | 35 | -20 | 27 | 36 | 8 | 0 | 11 |
| 11 | 71 | 69 | 20 | 4 | 91 | 7 | 0 | 10,16 |
| 12 | 16 | 26 | -20 | 2 | 153 | 4 | 0 | 8,13 |
| 13 | 85 | 56 | 20 | 92 | 182 | 2 | 12 | 0 |
| 14 | 57 | 37 | 20 | 47 | 126 | 19 | 15 | 0 |
| 15 | 61 | 96 | -20 | 69 | 112 | 1 | 0 | 14 |
| 16 | 12 | 17 | -20 | 35 | 94 | 18 | 0 | 11 |

Le tableau 2 illustre le résultat de la simulation qui représente la solution qui minimise la somme pondérée des trois critères.

**Tableau 2 : Résultat de la simulation**

| M in $f$ | Chemin parcouru |
|---|---|
| 3911.339 | 0 *4 3 2 1 4 7 5 9  1 8 12 6  11 16 11 10 14 15 13 12* 0 |

La fonction $f$ est exprimée en unité de coût.

Le chemin parcouru indique l'utilisation d'un seul véhicule permettant de minimiser la fonction coût. Notant que dans notre application, le nombre de véhicules utilisés varie entre 1 et K. On remarque ainsi qu'un nœud peut être visité plusieurs fois par le même véhicule, le cas des nœuds 1, 4, 11 et 12.

## 6. Conclusion

Nous avons proposé, suite à une présentation de l'état de l'art du VRP et une étude des différentes approches proposées pour la résolution du 1-PDPTW et du m-PDPTW, une approche multicritère basée sur les algorithmes génétiques pour la minimisation de la somme pondérée du nombre de véhicules utilisés, la somme des retards et le coût total de transport, et ce en se basant sur la méthode de Pareto optimalité.